\documentclass[runningheads]{llncs}
\usepackage{graphicx} % Required for inserting images
\usepackage{subfigure}
\usepackage{subfig}
\usepackage{tabularx}
\usepackage{cite}

\title{Learning Run-time Safety Monitors for Machine Learning components}
\author{Ozan Vardal\inst{1}  \and Richard Hawkins\inst{1} \and Colin Paterson\inst{1} \and Chiara Picardi\inst{1} \and Daniel Omeiza\inst{2} \and Lars Kunze\inst{2} \and Ibrahim Habli\inst{1}}

\institute{Department of Computer Science, University of York, York, UK \email{\{ozan.vardal,richard.hawkins,colin.paterson\}@york.ac.uk} \and Department of Engineering, University of Oxford, Oxford, UK}

\authorrunning{Vardal et al.}

\begin{document}

\maketitle

\begin{abstract}
For machine learning components used as part of autonomous systems (AS) in carrying out critical tasks it is crucial that assurance of the models can be maintained in the face of post-deployment changes (such as changes in the operating environment of the system). A critical part of this is to be able to monitor when the performance of the model at runtime (as a result of changes) poses a safety risk to the system. This is a particularly difficult challenge when ground truth is unavailable at runtime. In this paper we introduce a process for creating safety monitors for ML components through the use of degraded datasets and machine learning. The safety monitor that is created is deployed to the AS in parallel to the ML component to provide a prediction
of the safety risk associated with the model output. We demonstrate the viability of our approach through some initial experiments using publicly available speed sign datasets.

\end{abstract}

%%Research highlights
%\begin{highlights}
%\item Research highlight 1
%\item Research highlight 2
%\end{highlights}

\section{Introduction}
\label{sec:intro}

\begin{figure*}
\includegraphics[width=1\textwidth]{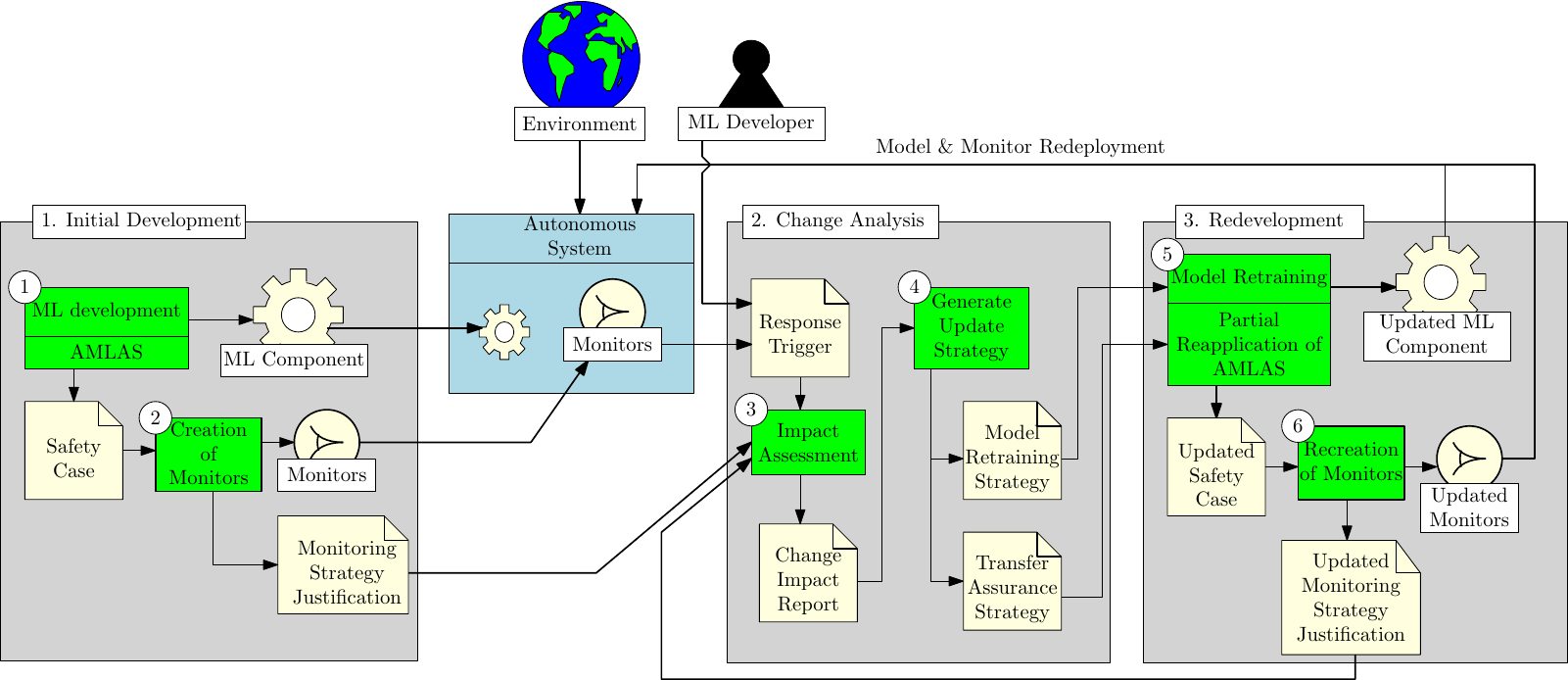}
\caption{Three Stage Transfer Assurance Process with activities shown in green and artefacts in yellow.}
\label{fig:trAss}
\end{figure*}

The use of machine learning (ML) in perception and understanding is essential for many autonomous systems (AS). Where such systems are used for safety related tasks it is critical that the safety of the ML components can be assured prior to deployment. Post deployment, we must be able to demonstrate that the system continues to operatate safely throughout operation in complex and dynamic environments. In \cite{picardi2023transfer} we introduced transfer assurance, a process for assuring ML components used in AS. Transfer assurance is used  when the ML component is required to be updated in response to changes in the AS or, crucially, in response to changes in the environment in which it operates. 
%The transfer assurance process aims to ensure that the safety of the ML component continues to be assured in the presence of such post-deployment changes. 
Figure \ref{fig:trAss} shows an overview of our transfer assurance process, which is split into three stages containing six activities. The first stage considers the initial development of the ML component for deployment into an operational AS and makes use of the AMLAS assurance process~\cite{hawkins2021guidance}. This stage results in the creation of an ML component along with its safety case and a set of appropriate ML safety monitors. These ML safety monitors are deployed on the AS along with the ML component and are used to identify changes in the system or environment which invalidate the safety case for the ML component. The second and the third phases deal with analysing and responding to such changes to maintain acceptable safety of the AS. This paper focuses on the first stage of the transfer assurance process, in particular the creation of effective safety monitors which allow us to understand the impact of change on ML models at run-time (activities 2 and 3 in Figure \ref{fig:trAss}).

The challenge for activity 2 is to create monitors capable of detecting when ML component outputs are potentially unsafe due to changes in operating conditions.
This is particularly challenging because we are unable to determine ground-truth during operation and because changes in operating conditions are combinatorial in nature, meaning that only by understanding the complex interplay of features can we determine the impact on safety. 

We address this challenge by training an ML safety monitor with degraded data to reflect the potential impact of real-world  factors. 
These degraded data sets are presented to the ML component and labelled to reflect their impact on model performance. 
This labelled data is then used as training data to learn a new model that can be used as a safety monitor during operation of the AS. 
This approach enables indication of the level of safety risk for the model given current input conditions. 

This safety monitor is then deployed to the AS alongside the ML component, taking the same inputs from the operating environment at run-time and providing a prediction of the safety risk associated with the model output (activity 3), potentially triggering a response where the safety risk is assessed to be significant. The response may involve action from the system to ensure the continued safe operation of the AS (for example through fall-back to alternate systems or restricted operation) as well as consideration of the impact on the safety case.

\iffalse
\begin{figure*}
\includegraphics[width=1.0\textwidth]{images/monitorRuntimev2.pdf}
\caption{Deployment architecture for monitor\label{fig:runtime}}
\end{figure*}
\fi

The rest of the paper is organised as follows.
In Section \ref{sec:related} we start by discussing related work before moving on to describe our proposed methodology in Section \ref{sec:meth}.
Next, in Section \ref{sec:evaluation}, we describe experiments we have undertaken to demonstrate our approach before finally, we discuss future work and provide conclusions in Sections \ref{sec:disc} and \ref{sec:conc}.

\iffalse
, causing an uncertain prediction which decreases the assessed performance of the model and invalidates its assurance.  In the process described in our previous work \cite{picardi2023transfer} the monitor is accompanied by a document called “Monitoring Strategy Justification” which contains all the rationale of the monitor: what are the variables monitored, why (e.g. how they affect both the model and its assurance) and how (e.g. technology involved).  This document plays a key function in assessing the impact of the changes on both the model and its assurance case.
The impact assessment involves the human analysis of the response trigger combined with the Monitoring Strategy Justification document in order to assess how the change impacted the model and its associated assurance case. We address this in the paper discussing the rationale of the process after a possible response trigger.
In the following sections we will explain our methodology to create the monitor illustrating the case of study, the model and the dataset used. The paper is organised as follows:....
\fi

\subsection{Related work}
\label{sec:related}

This work fits in the broader research area of monitoring ML models during operation through the detection of \textit{data distribution shift}, which refers to the phenomenon in supervised learning where data in operation change after deployment, differing from the training data used to develop the model \cite{huyen2022designing}.
This misalignment between the training and testing distribution can cause models to predict less accurately, despite maintaining confidence that their performance is good \cite{Amodei2016, Rabanser2019}.
This is a significant safety issue, as erroneous predictions that are not appropriately flagged for inspection can lead to harmful outcomes.
For example, changes in medical imaging software or hardware that are not anticipated by an ML component could endanger patients through over- or underdiagnosis \cite{Finlayson2021, Roschewitz2023}.

Detecting samples that are out-of-distribution (OOD) is a problem that has received much research attention in recent years \cite{Hendrycks2016, Liu2020, Shafaei2018, Hashemi2023}.
Hendrycks et al. introduced a solution for detecting OOD samples in neural networks by thresholding softmax probabilities \cite{Hendrycks2016}.
More recently, Liu et al. proposed use of energy scores, derived from energy-based models, as a more effective method than softmax confidence scores \cite{Liu2020}.
Another promising approach is use of non-parametric and parametric statistical tests for multiple conditional distribution hypothesis testing \cite{Kulinski2021}.
This approach can can pinpoint the exact features causing data distribution shifts, addressing shortcomings of traditional OOD detection methods.
Other approaches involve use of general adversarial networks (GANs) to detect anomalies \cite{Deecke2018, Carrara2021, Perera2019} in high-dimensional data.

Effective safety monitoring may require more than identifying OOD inputs, as not all OOD data pose safety risks, and recent work has demonstrated that performant OOD detection does not necessarily translate to effective safety monitoring in practical scenarios \cite{Guerin2023}.
Our work differs from OOD detection as we propose an approach to identifying the boundaries at which degradation of input data may make outputs from the ML component poze a hazard for the AS.
Our primary contribution is the introduction of a process explicitly designed to create safety monitors for ML components used in complex environments.
We demonstrate the effectiveness of our proposed process through application to an image classification problem for sign identification in an autonomous vehicle.
\section{Methodology}
\label{sec:meth}

In this section we describe the methodology we have developed for the creation and use of safety monitors that identify when changes to the post-deployment operational environment of the AS have the potential to invalidate the safety case for the ML component.

\begin{figure*}[!t]
    \includegraphics[width=1\textwidth]{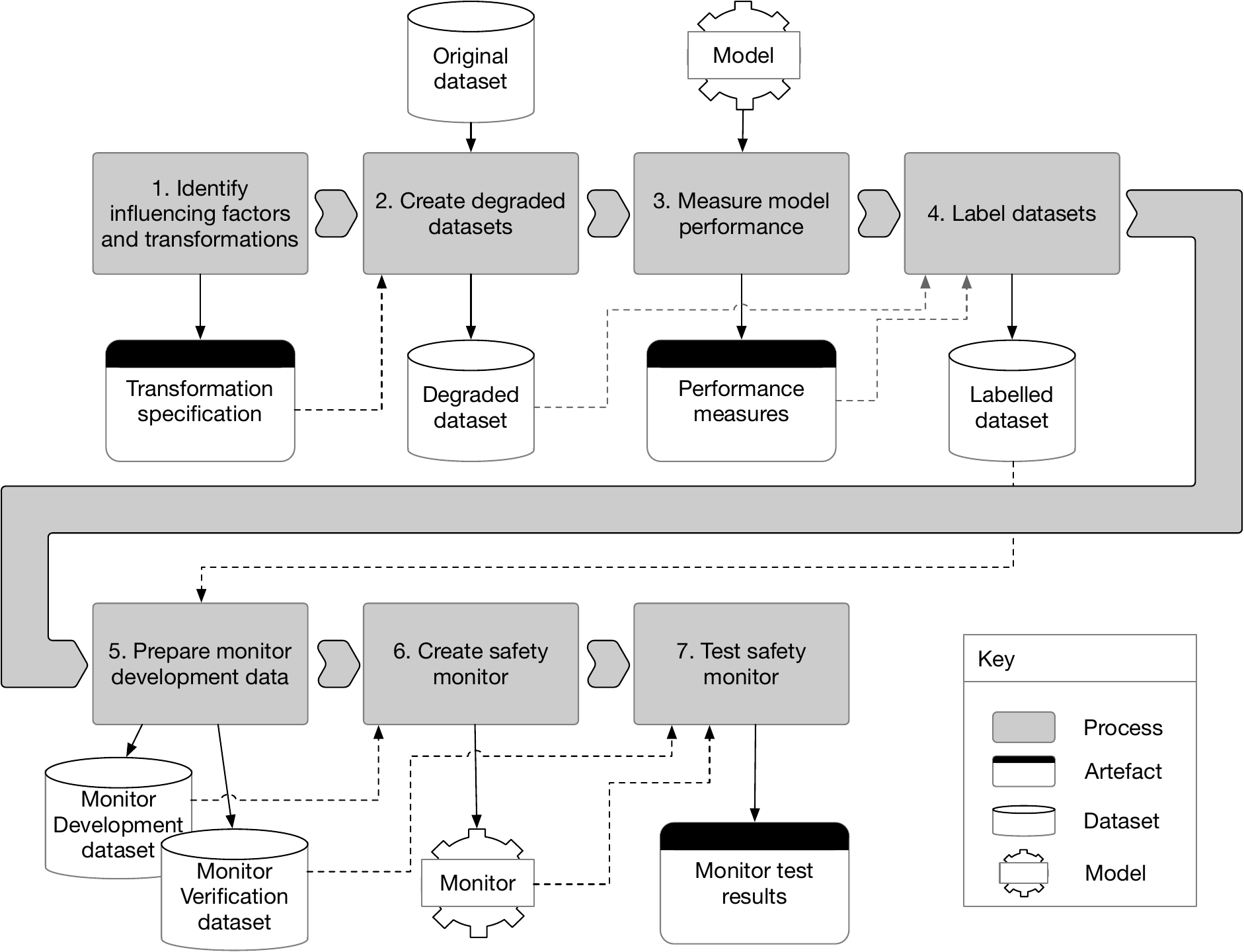}
    \caption{Overview of process for creation of ML safety monitors }
    \label{fig:process}
\end{figure*}

The seven step methodology, shown in Figure \ref{fig:process},  starts with the identification of influencing factors that are likely to impact the performance the ML component and the safety of the AS. These influencing factors, e.g. blur and noise, arise from the combinatorial effects of natural phenomena in the operating environment and the operation of the sensors used in the AS.
Once these have been identified, transformations must be specified to simulate the effects of these influencing factors on unperturbed input data (Step 1).
For instance, approaches to such transformations could include mathematical functions that transform input data to produce perturbed output, or other data augmentation methods such as GANs \cite{Tanaka2019} used to simulate input data perturbed by influencing factors in a realistic fashion.

The next step is to identify an unperturbed dataset that is representative of input data likely to occur in the operating environment of the AS, and systematically apply the defined transformations to this dataset at different levels of degradation to create a set of degraded datasets (Step 2).

Once a degraded dataset of each combination of influencing factors and perturbation levels has been created, the performance of the ML component on each degraded dataset is recorded (Step 3).

Based on these observations, each dataset is labelled according to its correspondence to a pre-determined performance threshold (Step 4). These thresholds are derived through an assessment of the impact of the performance of the ML component on the safety of the AS, informed by system safety analysis and aligned with levels of safety risk identified in the AS safety case. 
As an example, based on a consideration of required system behaviour, a dataset may belong to a  ``safety level 1: Normal operation" if the performance of the ML component on that dataset is above 90\%, or "safety level 3: Emergency stop" if the performance is  below 20\%.

Labelled datasets are then aggregated and split into training and development data in preparation for a typical machine learning training procedure (Step 5), whereby the safety monitor model is trained to classify samples as belonging to different levels of safety (Step 6).
%, with classes representing the level of safety as determined by the ML component's performance on the degraded datasets. 
This classification allows the safety monitor to estimate the operational safety of the ML component used in an AS in real-world conditions by identifying the impact of natural phenomena in the operating environment on the images being presented to the ML models. 

%comparing incoming data against the trained model's understanding of different levels of degradation and their impact on performance. 

Finally in (Step 7), the performance of the safety monitor is evaluated using established validation methods. 
This can include testing on external datasets for unbiased assessment or using k-fold cross-validation techniques to ensure robustness and validity.
%We believe that the methodology is a generalisable approach for the creation of safety monitors for use in a variety of contexts.
%we present what we believe to be a generalisable approach to creating ML safety monitors that can be realised in a variety of ways and contexts.
%However, we acknowledge that the manner in which any given step is realised will depend on the complexity and nuances of the operating domain of the AS for which the safety monitor is being trained.

In order to illustrate our methodology, we demonstrate how it may be applied to an image classification task in an automotive context. 
We considered a safety monitor trained to work alongside a road sign classifier. 
The ML component that the safety monitor was trained for was a pre-trained convolutional model that had been developed for a previous study (see model 3b in \cite{paterson2021deepcert}). 

In the following we explain the rationale underlying each step of our methodology in more detail.

\vspace{0.5cm}
\noindent
\textbf{Step 1: Identifying influencing factors and transformations:} The first stage of the safety monitor training process involves identification of \textit{influencing factors} that affect the ML component during operation in ways that may compromise its safety.
These factors can be identified through a consideration of internal factors, i.e. system features, and external factors, i.e. natural phenomena in the operating environment.
%Internal factors can include degradation in sensor performance. External factors include the operating environment of the AS such as effects of weather or occlusions in the real-world.

%For initial validation, we present a simple example for demonstrative purposes. 
%In real-world contexts, it will likely be necessary to use more sophisticated approaches in light of the complexity of a typical autonomous systems, for instance...

%While these factors are highly context-specific, we consider an autonomous vehicle application in which a neural network responsible for road sign classification may encounter environmental challenges.
For our example application, the operating environment of the ML component is the road network on which the vehicle is operating, and the input data provided to the ML component are images of road signs captured by the vehicle's camera sensors.
%The way in which the AS captures and provides these input data to the ML component is through the vehicle's image sensors, and the activities performed by the ML component on the input data are the classification of road signs.
We therefore identify two factors that may influence the performance of the ML component: weather conditions such as fog and heavy rain.
These factors were chosen as we anticipate that elevated levels in some or all of these factors can lead to to misidentification of road signs.
This may affect the operating safety of the AS, for example, by resulting in incorrect speed adjustments which place the vehicle in q hazardous state.

%To embed these influencing factors into the safety monitor, appropriate transformations must be identified to simulate the effects of influencing factors on the input data.
Rather than attempting to model all possible real-world phenomena we consider the effect of these, in combination, on the image as presented to the ML model. For example rain, fog and particulates may all be present causing colour shifts, blur and noise in the image. It is not important for the safety monitor to understand the source of these image perturbations, but rather the combined effect of these on the image and the potential impact on safety.

These transformations may take the form of mathematical functions which take an image sample as input and produce a degraded image that mirrors the effects of the identified influencing factors on the input data.
Building on our previous work on test-based verification of neural network image classifiers \cite{paterson2021deepcert}, we define a perturbation encoding function $g$ as transforming an image $X$ into a set $Z$ consisting of a single perturbed image $X'$, achieved by the application of a perturbation function on a per-pixel basis:
\begin{equation}
x'_{i,j} = \textit{perturbation}(X_{i,j}, \epsilon),
\end{equation}
where $x'_{i,j}$ represents the pixel at coordinates $(i, j)$ in the perturbed image $X'$, and $X_{i,j}$ is the corresponding subset of pixels from the original image $X$, and where $\epsilon \in [0, 1]$ serves as a measure of the intensity of perturbation.
For images with color, the image $X$ is encoded as an array of pixels, with each pixel containing a triplet of values that specify the color's red, green, and blue components.

Having defined the general mathematical form of the transformation in our image classification context, we then consider how it may be adapted for each influencing factor we anticipate to affect the ML component during operation.
For example, to simulate the effects of haze in the image, due to fog or water spray, we can apply a \textit{haze transformation} to each image, which may be defined as a color overlay that perturbes the image.
Assuming a uniform haze effect, a haze color can be characterized by $C_f = (r, g, b)$. This effect is integrated into the image via the equation:

\begin{equation}
x'_{i,j} = (1 - \epsilon)x_{i,j} + \epsilon C^f,
\end{equation}

A value of $\epsilon = 0$ signifies no alteration to the image, whereas $\epsilon = 1$ results in the image being completely filled with the color $C^f$.
Operations on the pixel values are conducted element-wise.

To simulate blur, which may arise through rain or grease on the camera lens, we can perturb each image using a \textit{blur transformation}.
This is modelled using a convolutional filter with a Gaussian kernel, where each pixel $x'_{i,j}$ in the blurred image is a weighted average of its neighbors:

\begin{equation}
x'_{i,j} = \sum_{k=-k_d}^{k_d} \sum_{l=-k_d}^{k_d} \alpha_{k,l} \cdot x_{i+k,j+l},
\end{equation}
with weights $\alpha_{k,l} \in [0, 1]$ derived from a discretized Gaussian distribution, normalized such that $\sum_{k=-k_d}^{k_d} \sum_{l=-k_d}^{k_d} \alpha_{k,l} = 1$.
The perturbation intensity $\epsilon$ scales with the Gaussian standard deviation, dictating the blur level.

\iffalse
% experiments that include all 3 factors are memory intensive and require some optimisation to run
Finally, we may choose to simulate the effects of sensor degradation by applying a \textit{salt and pepper noise} transformation to each image, with "salt" representing hot pixels that appear white, and "pepper" representing permanently damaged pixels that receive no input, thus appearing black.
The quantity of pixels modified is a function of $\epsilon \times \text{total pixels in } X$.
Specifically, the process randomly assigns some pixels to "salt" (setting them to maximum intensity) and others to "pepper" (setting them to minimum intensity). For any pixel at location $(i,j)$:

\begin{equation}
X'_{i,j} = 
\begin{cases} 
0 & \text{if } (i,j) \text{ is selected for pepper noise} \\
1 & \text{if } (i,j) \text{ is selected for salt noise} \\
X_{i,j} & \text{otherwise}
\end{cases}
\end{equation}
\fi

\vspace{0.5cm}
\noindent
\textbf{Step 2: Creating degraded datasets:} 
In the next step, influencing factors and corresponding transformations identified in Step 1 are embedded into datasets used to train the safety monitor. 
This process has several substages: 

    1. An initial dataset that is maximally representative of an unperturbed operating environment must first be obtained. 
		For an image classifier, this could be a dataset of images captured in the environment that the ML component is expected to operate in.
		In the case of our road sign classification model, we use a subset of original images from the German Traffic Sign Recognition Benchmark (GTSRB) dataset.
  %(note that this dataset would us to train a safety monitor for exclusively for use on a german road network).
%		The choice of dataset is highly context specific, and while we demonstrate our methodology using static images in a vehicular context, non-static data may be more appropriate.
%		In an entirely different context, such as clinical decision support system that takes physiological data as input, data may be tabular in nature.

    2. Transformations specified in Step 1 must be suitably encoded so as to perturb any given individual image sample at a given value $\epsilon$. 
		Transformations may be applied uniformly across every point in a sample, or may be applied non-uniformly, for example, by applying a blur transformation to the center of an image, and a haze transformation to the edges. Indeed, the application methodology may be considered as an additional influencing factor, with the amount of image, or location, effected as a hyperparameter.
		For simplicity, in our example, we choose to apply each transformation uniformly to each pixel in the entire image.
  
	3. Ranges of perturbations to apply to each sample must be identified.
		This specification can take the form of a vector $E$ of intensity values $\epsilon$, where each $\epsilon$ represents the intensity of a particular transformation to be applied to each sample.
		It is important to consider upper and lower limits that would be appropriate in light of the anticipated safety domains in the operating environment of the ML component.
		For example, if the ML component is expected to operate in a domain where fog is omnipresent, then the lower limit of the fog perturbation should be set to a value that is representative of the minimum amount of fog that the ML component is expected to encounter.
		Similarly, the upper limit of the fog perturbation should be set to a value that is representative of the maximum amount of fog that the ML component is expected to encounter.
  
%        These limits can be set based
	4.  %, such as the presence of all three of rain, fog, and sensor degradation in our road sign classification example.
	 We also consider the possibility that multiple individual sources of degradation may interact combinatorially to impact the image classification capability of the ML component and aim for coverage in the combined space.
		%These interactions can be accounted for by considering the multiplicative combinatorial product of the ranges of perturbations identified in the previous step.
		We therefore generate a set of all possible combinations of perturbation values that can be applied to each sample.
		%For example, if we have three transformations, each with five possible perturbation values, then the combinatorial product would be a set of $5^3 = 125$ possible combinations of perturbation values.
		Each combination of perturbation values represents a degraded dataset.
  
	5. Once the above steps have been completed, each degraded dataset is created by sequentially applying each combination of transformations at perturbation values specified in the combinatorial product above to each sample in the initial dataset.
		For example, if we have two transformations, each with five possible perturbation values, and an initial dataset of 100 samples, then the creation of each degraded dataset would involve applying each of the 25 combinations of perturbation values to each of the 100 samples, resulting in a total of 25 degraded datasets, each comprising 100 samples with identical levels and types of perturbations within each dataset.

At the end of this process, a set of degraded datasets is generated that can represent the full spectrum of anticipated operating environments of the ML component, ranging from mild to severe degradation.

%\textcolor{red}{Need to say somewhere how we justify that we have the right combination of transformations}
%\textcolor{red}{Also, do we need to say what the transformation factors might be for non-image based datasets?}

% \begin{figure*}
% \includegraphics[width=1.0\textwidth]{images/datasetCreation.pdf}
% \caption{Process to produce a new dataset for the monitor creation.\label{fig:data}}
% \end{figure*}

\vspace{0.5cm}
\noindent
\textbf{Step 3: Measuring model performance:}
The next step in our methodology involves quantitatively assessing how environmental changes impact the ML component's performance.
This assessment is critical for understanding and anticipating the conditions under which the ML component's reliability may waver.
Once the degraded datasets have been created, they are provided as input to the ML component that a safety monitor is being developed for, and the updated performance of the model on each dataset ($p_1,  p_2, \ldots,  p_m$) is observed.
%These performance measures can be single value or a vector.

For our example we make use of the image classification model used in automotive scenarios taken from our earlier work~\cite{paterson2021deepcert}.
We evaluate model performance across our range of contextually relevant degradations by using each degraded dataset as a test set for the original model and recording its accuracy, thereby establishing a performance baseline under various distortion conditions. Although we only record accuracy here, multiple performance metrics may also be considered.

\vspace{0.5cm}
\noindent
\textbf{Step 4: Labelling datasets:}  
Our methodology assumes that there exists an acceptably safe state  for the system which is related to the current model performance. Indeed this should be argued prior to deployment in the system safety case~\cite{hawkins2021guidance}.
Distinguishing between acceptable and unacceptable operating states requires a specification of performance boundaries at which the system may move between states.

It is therefore necessary in our approach to first specify a vector of performance thresholds that define the boundaries of acceptable performance for the model, and then creating labels $l_1,  l_2, \ldots,  l_m$ for each degraded dataset to represent the expected operating state of the model based on its performance in a degraded environment.
Thus, $l_1$ would indicate very degraded accuracy and a hazardous operating state, whereas $l_m$ would indicate almost the same accuracy as achieved on the original dataset and an acceptable operating state.

For our example case we assume three modes of operation exist. When the performance of the model (accuracy) is over 70\% then the vehicle may operate normally. Between 70\% and 40\%, the vehicle would have a limited top speed giving the decision making component more time to gather data from the image sensor and hence mitigate the increased uncertainty in the input stream. Finally, when accuracy drops below 40\% the vehicle is unable to continue to operate safely and a graceful cessation of autonomous function is initiated, this might be handing off to a human operative or bringing the vehicle to a halt. For such a case the thresholds are defined as $[70, 40]$.

%For instance, in our specific case with the traffic sign recognition model, we implement this approach by setting distinct accuracy thresholds $[70, 40]$ that reflect the model's ability to cope with our anticipated environmental degradations.
%The degraded datasets from the previous step are then classified into three major categories based on the model's accuracy: the first category includes datasets where the model's accuracy remains above 70\%, indicating a relatively minor impact on performance.
%The second category encompasses datasets where the model's accuracy falls between 40\% and 70\%, signaling a moderate degradation in performance that could potentially affect the model's safety.
%The final category is reserved for those instances where the model's accuracy dips below 40\%, representing significant degradation likely to severely compromise the model's safety in real-world scenarios.

The datasets are then labelled according to these thresholds and the labeling forms the basis for the safety monitor's training. In this way we ensure that the monitor is attuned to detect and respond to performance dips analogous to those it might encounter during the autonomous system's operation.

\vspace{0.5cm}
\noindent
\textbf{Step 5: Preparing monitor development data:}
The creation of an effective safety monitor relies on the preparation of appropriate training data.
This process adopted for safety monitor creation follows a typical machine learning process \cite{Raschka2018, huyen2022designing}: labelled datasets are divided into development data and testing data.
The development data are used for training and validation of the safety monitor. 
The test data is then used to evaluate the generalisability of the created safety monitor by checking that the safety monitor generates correct labels for the testing data. 

In our application with the traffic sign recognition model, this translates into assembling a dataset that encompasses a range of degraded images, each labeled according to the previously established accuracy categories.
In our example, we chose an 80:20 split of training to validation data, as this is common practice in most ML training contexts \cite{Joseph2022}.

\vspace{0.5cm}
\noindent
\textbf{Step 6: Creating the monitor:}
The safety monitor is an ML model that is created by following the traditional ML cycle of training and testing~\cite{ashmore2021assuring}.
%During this process some decisions like early stopping criteria or method of splitting data into development and training data need to be taken.
%These decisions will ultimately vary based on the specific context of the safety monitor and the ML component it is being developed for.
For our case, the safety monitor for the traffic sign recognition model is a convolutional neural network trained to classify images based on the performance labels $l_1,  l_2, \ldots,  l_m$ specified in steps 4 and 5.
Note that this structure was not extensively tuned and is purely to demonstrate the application of our methodology.
The safety monitor is then trained to detect the severity of performance degradation in the model based on fluctuations in haze and blur.%, and slat and pepper noise.

\vspace{0.5cm}
\noindent
\textbf{Step 7: Testing the monitor:}
The final step involves evaluating the safety monitor's performance on a set of test data similarly degraded.
For our example case, we tested the resultant safety monitor using a 5-fold cross-validation procedure.
\section{Evaluation}
\label{sec:evaluation}
In this section we evaluate the use of our methodology and illustrate the application of each stage as described in the previous section.

We make use of a subset of the German Traffic sign benchmark \cite{stallkamp2011german} where each sample represents a speed sign as a 32x32 pixel RGB image. 
The ML component, for which we are designing a monitor, has a nominal accuracy, prior to image degredation, of 0.98 when assessed with its associated test set.
This test set contains 4110 images and is described in Table \ref{table:data} where the number of samples for each of the classes are reported.

%Following Step 1 of our methodology, we identified fog and rain as influencing factors that have the potential to compromise the performance, and thus safety, of our road sign classifier.
%We then specified mathematical transformations to simulate the effects of these influencing factors on the image data. 
%We used a haze transformation to represent fog, applying a uniform color overlay to reduce visibility in the images.
%For rain, we simulated water droplets obscuring parts of the image through a blur transformation using a Gaussian kernel to mimic the blurring effect of water on the lens.
%These are described in more detail in the methodology section above.

In step 1 of the methodology we applied the blur and haze transformation, as previously described, to the original, unperturbed, dataset. We uniformly varied the intensity of these factors for each image.
Initially, five different perturbation levels were applied for each factor: $E = [0, 0.2, 0.5, 0.8, 1]$.
%Our combination of five perturbation levels and three influencing factors resulted in a combinatorial product of 125 degraded datasets, where each degraded dataset was a result of each unique combination of perturbation levels across the three factors.
The number of degraded images may be calculated as
\begin{equation}
N = n_{i}  \times \rho^{n_f}
\end{equation}
where $\rho$ is the number perturbation levels in the set $E$, $n_i$ is the number of initial images to which the degradation will be applied and $n_f$ is the number of factors to be applied. For our case of $n_i = 4110$, $n_f=2$ and $\rho = 5$ we generate 102,750 degraded images.

%Each of these degraded datasets, like the original dataset, contains 4110 images that have been degraded according to the respective perturbation levels ($\epsilon$ values) for that run of degradation.

\begin{figure*}
    \centering
    \includegraphics[width=1\linewidth]{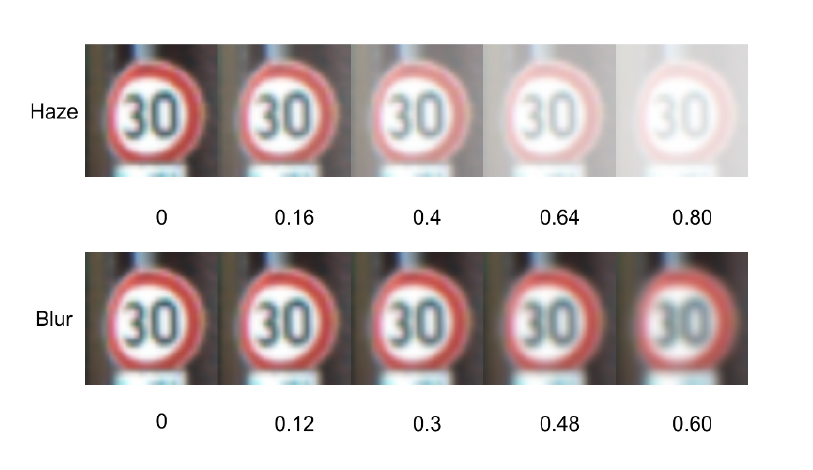}
    \caption{Perturbed samples from the GTSRB dataset showing the independent effects of the influencing factors. Numbers under the images indicate the level of the effect applied.}
    \label{fig:perturbed_samples}
\end{figure*}

Next we inspected the generated images to assess the range of perturbation for each factor. This is to ensure that we are not generating unrealistic images. We then redefined our perturbation range such that the set of epsilon values for haze, and blur where $E_{haze} = 0.8*[0, 0.2, 0.5, 0.8, 1]$ and  $E_{blur} = 0.6*[0, 0.2, 0.5, 0.8, 1]$, respectively.
Examples of unperturbed and degraded images, together with corresponding perturbation levels, are provided in Figure \ref{fig:perturbed_samples}.
While this is a simple demonstration of how synthetic transformations can be applied and tuned for a given context, we acknowledge that real-world examples are complex and likely warrant more a larger number of factors and associated mathematical transforms.

\begin{table}[!t]
    \begin{center}
        \begin{tabular}{|c|c|c|c|c|c|c|c|} \hline
        \textbf{class} & 0 & 1 & 2 & 3 & 4 & 5 & 6\\ \hline
        \textbf{description} & 30 km/h & 50 km/h & 60 km/h & 70 km/h & 80 km/h & 100 km/h & 120 km/h \\ \hline
        \textbf{\# data samples} & 720& 750 & 450 & 660 & 630 & 450 & 450 \\ \hline
        \end{tabular}
    \end{center}
    \caption{Original test dataset with class descriptions and sample counts.}
    \label{table:data}
\end{table}

\begin{figure*}[]
	\centering
    \includegraphics[width=1\textwidth]{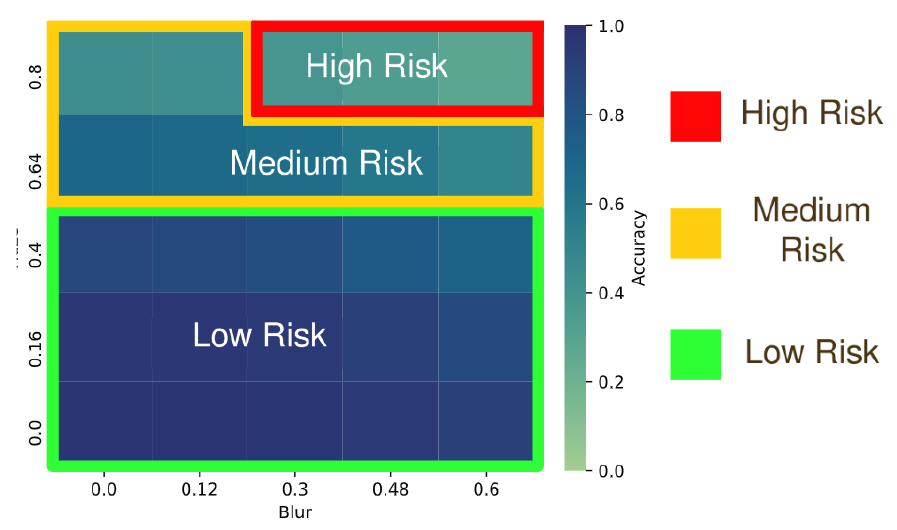}
    \caption{Heatmap of accuracy of the original road sign classifier for each degraded dataset. Axes represent the $\epsilon$ (degradation level) of each degradation effect. The region with the green border represents epsilon combinations that result in $accuracy \geq70\%$, while those with the amber and red border represent epsilon combinations that result in $40\%\leq accuracy<70\%$ and $accuracy<40\%$ respectively.
    }
    \label{fig:heatmap}
\end{figure*}

Each degraded dataset was then provided as input to the ML component and the accuracy of the model observed (Step 3).
For  the 25 degraded datasets generated  the model performed with accuracy $\geq$ 70\% for 15 datasets,  with 70\% $\geq$ accuracy $\geq$ 40\% on 7 datasets, and with accuracy $\geq$ 40\% on 3 datasets (see Figure \ref{fig:heatmap}). Labels were then addedd to each of the datasets (Step 4).
%The degraded datasets were then labelled (Step 4) according to the three classes described in table \ref{table:label}.
For larger numbers of factors a visualisation of this type would not be possible, however, the approach would still identify regions in the factor space which are mapped to modes of safe operation.

Labelled degraded datasets were then prepared as training data for a safety monitor tasked with classifying degraded datasets based on their performance label (Step 5).
Datasets for each class were concatenated and samples were randomly shuffled within each class.
Each concatenated dataset was then split into training and testing sets with an 80:20 split.
Once degraded datasets were shuffled, concatenated, and split into training and test sets, all training sets and test sets were concatenated.
The prepared data were then saved together with labels for classes in the training (82200 samples) and test set (20550 samples).

In Step 6, we trained the safety monitor by initializing a convolutional neural network and training it to distinguish between degraded samples belonging to different classes. To ensure that the safety monitor's performance was reliably evaluated we used 5-Fold cross-validation (this number of folds is standard practice \cite{Nti2021}), splitting each data set into 5 subsets, with each subset serving as a test set in one of the folds and as part of the training set in the others.

\begin{figure*}[]
	\centering
    \begin{subfigure}
        \centering
        \includegraphics[width=0.48\textwidth]{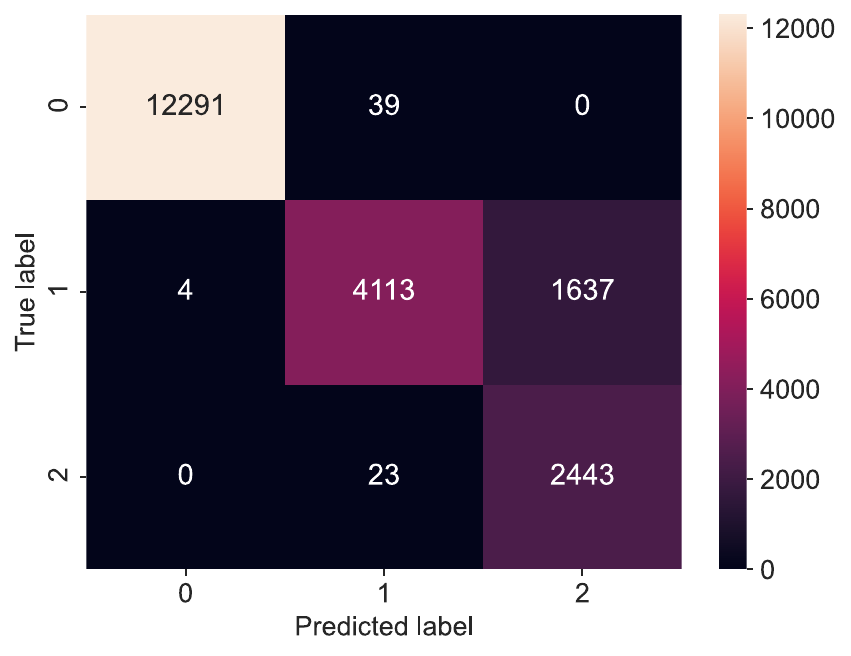}
        \label{fig:confusion_matrix}
    \end{subfigure}
    \begin{subfigure}
        \centering
        \includegraphics[width=0.5\textwidth]{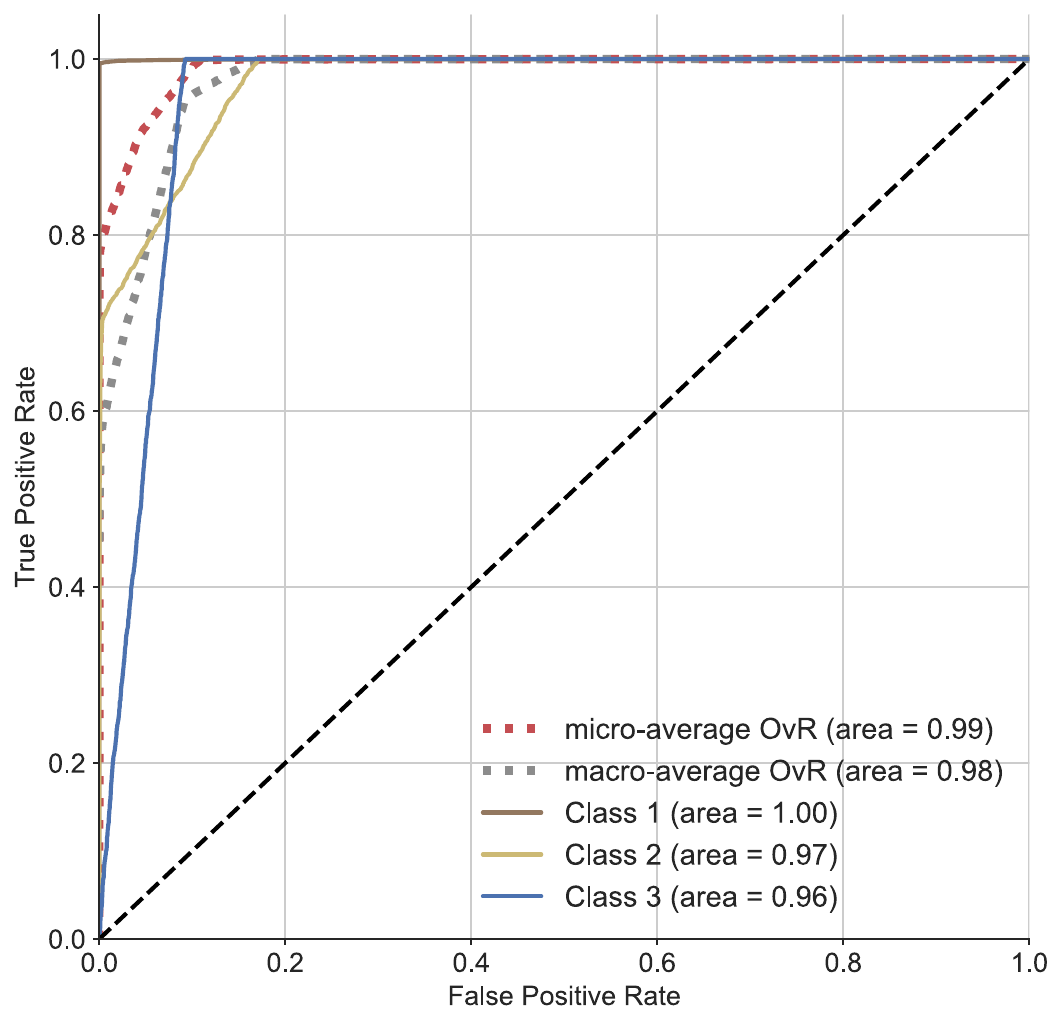}
        \label{fig:roc_curve}
    \end{subfigure}
    
    \caption{The left panel shows the performance of the safety monitor across three classes, with each class representing the anticipated operational safety of the ML component in environments with different levels of perturbation. The right panel shows Receiver Operating Characteristic (ROC) curves for each of the three operational safety labels classified by the safety monitor.}
    \label{fig:combined_results}
\end{figure*}

Finally, we evaluated the performance of the trained safety monitor by providing samples from the test set as input and measuring performance.
The final safety monitor model achieved an accuracy of 92\%.
Figure \ref{fig:combined_results} provides a confusion matrix of the monitor as well as aggregate and individual ROC curves for each class.

%The results are reported in table \ref{table:results}. 

\section{Discussion}
\label{sec:disc}

%\begin{itemize}
%    \item Extending the work to non-static data
%    \item Monitors for non-image data (e.g. tabular medical data)
%    \item data sets with greater complexity
%    \item safety monitor producing too many false positives may lead to human ignoring the alert (find citation for this)
%\end{itemize}

In order to illustrate our approach we have used a simple example based on a publicly available image dataset. While results of this case study demonstrate the feasibility of the approach, it is not limited in its application to a specific type of data or domain; our approach could in principle be used to create safety monitors for any ML component used in an AS. Further work will focus on testing the general applicability of the approach, with particular consideration for case studies involving non-image datasets. For example we are keen to test our approach to create safety monitors for ML components used in medical applications where the data takes the form of multi-attribute tabular patient data \cite{Jia2021}. One of the key challenges here, compared to the case study presented in this paper, will be the correct identification of relevant influencing factors and the different types of perturbations that will need to be considered.

The labelling of datasets used to train the safety monitor is crucial to the success of our process. For example if the risk regions defined for our use case in Figure \ref{fig:heatmap} are defined differently, it could result in a higher number of high risk outputs being missed by the monitor, or conversely a higher number of false warnings being provided by the monitor. Although the former case would seem to be of most concern, a high number of false positives could result in operational issues for the AS (if the system takes action to reduce perceived increased operational risk), and ultimately in warnings from the safety monitor being ignored or the monitor being ``turned off''. Further work will explore the effect of labeling decisions on the performance of the safety monitor and further guidance developed.

The examples we have used to-date have considered the inputs to the model as single-shot. In reality many of the applications of ML in AS will involve components presented with continuous data streams where a single erroneous output may be insufficient to affect the behaviour of the AS. In such systems it may be possible to create safety monitors that require multiple high-risk inputs before a warning is provided. Such a solution could also help to mitigate system operational issues discussed above. 
\section{Conclusions}
\label{sec:conc}

In this paper we have introduced a method that begins to address the problem of how to monitor the safety of an ML model at runtime when ground truth is unavailable. Establishing a method to do this is crucial when the model is being used as part of an autonomous system to perform critical tasks. This forms part of a broader process for ensuring that assurance of the ML model is maintained in the face of post-deployment changes.

We have undertaken initial experiments to demonstrate the viability of our approach, with promising results. Further work is required to validate the approach on more realistic examples. In particular we will explore the nature of peturbations that are applied to training data to ensure that they are reflective of the real-world operational context in which the model is being used. This will, for example, explore how to best reflect the non-uniform nature of real-world perturbations in degraded datasets.
\section{Acknowledgements}

This work was supported by the Engineering and Physical Sciences Research Council through the RAILS project (EP/W011344/1) and the Assuring Autonomy International Programme, a partnership between Lloyd’s Register Foundation and the University of York.
\bibliographystyle{splncs04}
\bibliography{bibliography}

\end{document}